\documentclass[10pt,a4paper]{article}

\usepackage{lrec2006}
\usepackage{mdwlist}
\usepackage{graphicx}
\usepackage[small]{caption}

\title{Analysing Temporally Annotated Corpora with CAVaT}
\name{Leon Derczynski, Robert Gaizauskas}

\address{University of Sheffield \\
		211 Portobello, S1 4DP, UK \\
		L.Derczynski@dcs.shef.ac.uk, R.Gaizauskas@dcs.shef.ac.uk}

\abstract{ 
We present CAVaT, a tool that performs Corpus Analysis and Validation for TimeML. CAVaT is an open source, modular checking utility for statistical analysis of features specific to temporally-annotated natural language corpora. It provides reporting, highlights salient links between a variety of general and time-specific linguistic features, and also validates a temporal annotation to ensure that it is logically consistent and sufficiently annotated. Uniquely, CAVaT provides analysis specific to TimeML-annotated temporal information. TimeML\footnote{See www.TimeML.org. TimeML has recently become an ISO standard~\cite{ISO08}.} is a standard for annotating temporal information in natural language text.  In this paper, we present the reporting part of CAVaT, and then its error-checking ability, including the workings of several novel TimeML document verification methods. This is followed by the execution of some example tasks using the tool to show relations between times, events, signals and links. We also demonstrate inconsistencies in a TimeML corpus (TimeBank) that have been detected with CAVaT.}

\begin{document}

\maketitleabstract

\section{Introduction}
In essence, TimeML mandates the mark up of expressions referring to \emph{times}, expressions denoting \emph{events} and expressions \emph{signalling} temporal relations between times and events or events and events; it also allows \emph{links} to be added between entities, which are labelled with the temporal relation holding between them.

Existing TimeML tools can be divided into two categories: those which produce or alter mark-up, for example to assist annotation, and those that perform analysis. Only a few tools have as yet been developed for TimeML, mostly focusing on the annotation task, such as TTK~\cite{verhagen2008temporal}, which does not support analysis. From the second category, in the absence of other software, the TimeML-using community is restricted to generic XML analysis tools, such as Xaira~\cite{burnard2003xara} or LT-XML\footnote{From http://www.ltg.ed.ac.uk/software/ltxml}, as well as similar format-specific tools (TEI). These generic corpus tools are powerful applications, but require substantial effort to apply to TimeML data. 

We have constructed CAVaT (Corpus Analysis and Validation for TimeML) to process collections of temporally annotated documents. CAVaT's functionality is divided into two main parts; an integrated browsing and report generation system, and a modular extensible error checking and corpus validation framework.

In this paper, we first describe the technical aspects of the tool. We then present the reporting part of CAVaT, and then its error-checking ability, followed by the execution of some example tasks using the tool. We present an overview of the tool's operation and capabilities in Section~2. This includes details of the corpus loading and folding process (Section~2.1), report generation, and also a detailed explanation of the advanced validation modules that are included with CAVaT (Section~2.3). A brief syntax summary is presented in Section~3; the full guide is on the CAVaT website\footnote{Available at http://code.google.com/p/cavat/}. Next, in Section~4, we present sample queries and output. In Section~5, we show inconsistencies and observations in a TimeML corpus (TimeBank) that have been detected with CAVaT. Finally, Section~6 summarises the tool and discusses future work.

\section{Overview of the tool}
\label{overview}
CAVaT is an open source tool, constructed from a set of Python modules and a database. It uses NLTK\footnote{See http://www.nltk.org/} and MySQL\footnote{See http://www.mysql.com/}. The interface is a text-based interactive prompt, and all operations are performed with text commands. Command syntax strives to be simple, flexible and close to natural language. After loading and pre-processing a TimeML corpus, one can analyse it using built-in reporting functions, and perform data validation with one of many checking components.

\subsection{Preprocessing}
\label{preprocessing}
CAVaT can work on any TimeML-annotated corpus that is stored as a collection of uncompressed files in a single directory, by importing it to a set of database tables. The corpus is initially processed by an XML parser (using Python's \texttt{minidom} and \texttt{expat} implementations), which retrieves document level data as well as all temporally annotated information, and places it into a MySQL database. Temporally annotated data includes all TimeML tags and their attributes, as well any enclosed tokens for EVENT, SIGNAL and TIMEX3 tags.

\begin{figure}
\begin{center}
\caption{Example text and TimeML annotation.}
\label{fig:timemlExample}
\includegraphics[scale=0.65]{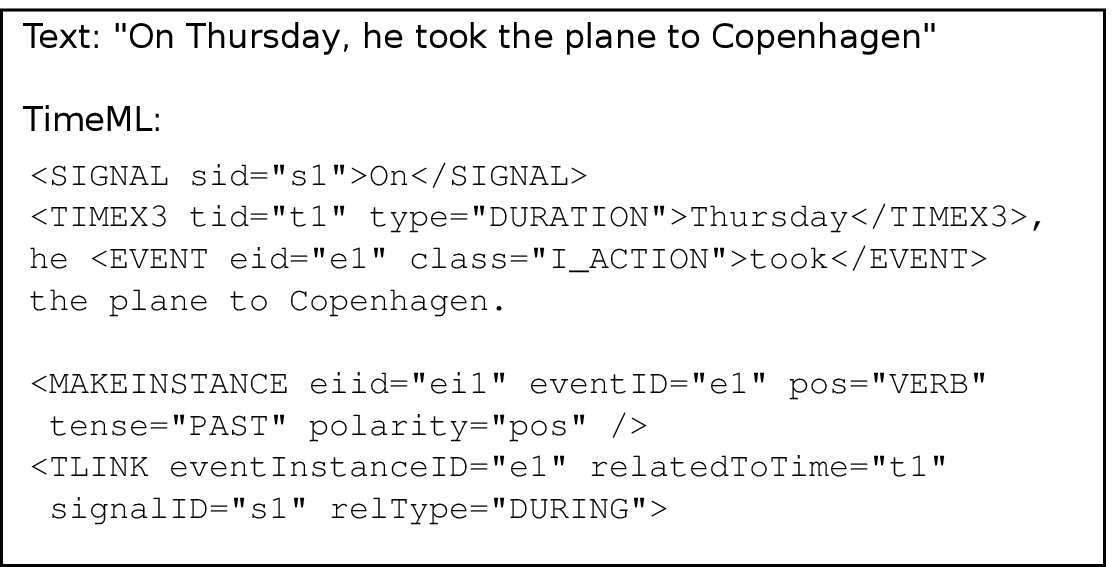}
\end{center}
\end{figure}

In TimeML, events are represented with the EVENT tag, and temporal expressions with the TIMEX3 tag. These intervals are the elements which CAVaT and the rest of this paper assume as temporal primitives, unless otherwise stated. Temporal relations between intervals are described with the TLINK tag, and temporal signals with SIGNAL. See Figure~\ref{fig:timemlExample} for an example.

Automatically classifying the type of temporal relation between intervals is currently a difficult problem in temporal processing of text~\cite{mani2006machine,lapata2006learning,hepple2007usfd}. The task is often made simpler by reducing the number of temporal link classes. TimeML includes \textsc{before} and \textsc{after} relations, though one may simply reverse the arguments of a \textsc{before} relation to turn it into an \textsc{after} one --- so, \emph{June 2008 was before August 2009} is equivalent to \emph{August 2009 was after June 2008}. It is thus possible to convert all links of one of these types to the other. We call this technique \textbf{folding}. Given a set of mappings, the 13 TimeML relations can be reduced.

\begin{table}
\begin{center}
\caption{Mappings between TimeML relations that can be applied in order to reduce the size of the relation set; when applying the transformation in the table, TLINK argument order is swapped.}
\label{tab:cavatFold}
\begin{tabular}{ | l | c | }
\hline
\textbf{Original relation type} & \textbf{Folds to relation} \\
\hline
\textsc{after} & \textsc{before} \\
\textsc{is\_included} & \textsc{includes} \\
\textsc{iafter} & \textsc{ibefore} \\
\textsc{begun\_by} & \textsc{begins} \\
\textsc{ended\_by} & \textsc{ends} \\
\textsc{during\_inv} & \textsc{simultaneous} \\
\textsc{during} & \textsc{simultaneous} \\
\textsc{simultaneous} & \textsc{simultaneous} \\
\hline
\end{tabular}
\end{center}
\end{table}

CAVaT offers three folds:
\begin{itemize}
\item \textbf{CAVaT fold} -- Collapses all inverse relations, such as mapping \textsc{included\_by} to \textsc{includes} (see Table~\ref{tab:cavatFold}).
\item \textbf{SputLink fold} -- The mapping introduced by Marc Verhagen, included in TTK~\cite{verhagen2005temporal}.
\item \textbf{Compact fold} -- Reduces TimeML's link relation set to 3 classes, using mappings defined in \newcite{setzer2005role}.
\end{itemize}
The first two are lossless, in that no temporal information is removed by the folding process. The third is lossy. It is possible to perform a lossy fold by, for example, reducing the TimeML \textsc{begun\_by} relationship to one of \textsc{includes}.

\subsection{Querying}
\label{querying}
The reporting part of CAVaT makes analysis of TimeML corpora simpler and easier than working directly with a set of XML documents, allowing flexible queries, and catering for inquiries specific to temporally-annotated data. 

The development of CAVaT has been driven by investigations of TimeML corpora. Many of the operations performed against a corpora had common elements, often centred around the retrieval of class distributions or token frequencies. A tool for TimeML corpus research could encompass all the required operations, while providing access to a larger range of reports.

CAVaT uses a report generation system where one can view any number of pre-defined features that match conditions of the user's choosing. Queries can produce reports at varying levels of granularity -- one may choose to examine data at sentence, document or corpus level.  Reports can output counts, distributions, lists or text extracts. Example queries are listed in Section~\ref{queries} Data such as part-of-speech information, tense, aspect, and event recurrence are captured by attributes described by TimeML, and any data like this (annotated by tags and their attributes) can be queried. In addition, properties specific to temporal data but not directly present in mark-up are implemented, including:

\begin{itemize}
\item \textbf{Event / event instance abstraction} In some cases, one piece of text may refer to two separate events (an example is given later in Section~\ref{tlink_loopCheck}). To permit annotation of this, TimeML's EVENT tag is placed around the text, and then event instances are specified using one or more MAKEINSTANCE tags. Data relating to a piece of event text, such as part of speech, polarity and modality, are described in the MAKEINSTANCE tag. However, we would often like to see the part of speech data for an event; indeed, when discussing temporal entities, the term ``event" is often used in place of ``event instance". Thus, CAVaT implicitly translates between these two related tags when requested; for example, when one asks to see event modality or cardinality.
\item \textbf{Signalled links} TLINKs may indicate a textual signal that suggests the type of relationship between their arguments. For example, in \emph{Lydia ate dinner before leaving the house}, the word \emph{before} acts as a signal, ordering two events. As signals are explicit indicators of temporal association, and correctly typing a temporal link is difficult, it is useful to be able to quickly identify which links employ a signal.
\item \textbf{Signal text and TLINKs} As SIGNAL text referenced from a TLINK may be thought of as that TLINK's signal text, CAVaT permits queries that specify signal text as an attribute of a TLINK.
\item \textbf{Text position and lemma} Although not part of the TimeML annotation schema, CAVaT logs text position (by sentence number and word number), and maintains lemmas of text found within tags.
\end{itemize}

One may view a particular TLINK's location in the original document, showing the link's arguments and their relation type. This helps understand the context of a single TLINK. For example, one may often see many links to a single document date, or discover that most links have arguments within the same paragraph -- something not immediately obvious to humans while browsing the TLINK markup, and unclear with generic corpus tools.

\subsection{Checking}
\label{checks}
Temporal annotation is a complex task, and as a result, a relatively small amount of text has been annotated to date. The largest TimeML corpus is TimeBank~\cite{pustejovsky2003timebank}, with less than 200 documents, and around 65000 tokens. Because of the complexity of temporal annotation, errors can arise beyond those that may be detected using an XML DTD. CAVaT is both a reporting and validation tool, and seeks to automatically detect high-level and complex errors that are rarely immediately obvious. Part of the motivation behind this part of the tool is similar to that of writing unit tests that highlight bugs in an application: to improve quality by automatically detecting previously seen errors. In this section we detail some checks that CAVaT can perform on a TimeML corpus.

Error checks are defined as Python modules, so that one may describe a detection method for an error case and share it with other researchers without modifying CAVaT's core code. The modules inherit from the \texttt{cavatModule} class; documentation is in the source code, and one may view a list of available modules with the command \texttt{check list}.

\subsubsection{Inconsistent closure}
\label{consistent}
\begin{figure}
\begin{center}
\caption{With time flowing from left to right, this represents $A$ \textsc{before} $B$ and $B$ \textsc{includes} $C$. It is not possible for $C$ and $A$ to be the same interval.}
\label{fig:intervals}
\includegraphics[scale=0.5]{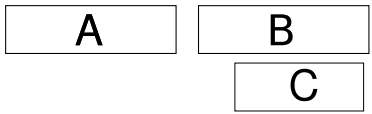}
\end{center}
\end{figure}

It is possible to create an inconsistent configuration of temporal links. For example, we may have $A$ \textsc{before} $B$ and $B$ \textsc{includes} $A$; this is clearly not possible, as \textsc{includes} stipulates that the start point of $A$ occur after the start point of $B$ (see Figure~\ref{fig:intervals}). While this example is fairly clear, it may not be at all clear to human annotators that a partial temporal link annotation could imply an inconsistent configuration. 

Automatically checking the consistency of a temporal network is hard. TimeML's relations are based on those of \newcite{allen1983maintaining}, and it is difficult to guarantee the consistency of networks formed using the latter set of relations~\cite{vilain1989constraint,tsang1987consistent}. We re-state the problem in a more simple fashion, as follows. Intervals are represented by pairs of endpoints, and we define intervals and the TimeML relations between them in terms of relations between these points. Our model uses only simultaneous ($=$) and before ($<$) relations.

The consistency checker works in a similar way to the closure algorithm in \newcite{setzer2005role}. It maintains an agenda and database. Assertions are taken from the agenda and used to infer further assertions when combined with assertions in the database. We initially process intervals in the document (taken from TLINK arguments) -- for each interval $I$ we add $I_{start} < I_{end}$ to the database. We then generate initial data for the agenda based on TLINKs in the document and a mapping for each TLINK to one or more assertions, listed in Table~\ref{tlink_assertions}.

The only inference rules needed with our minimal set of relations are:

\begin{itemize*}
\item If $x = y$ then $y = x$
\item If $x = y$ and $y = z$ then $x = z$
\item If $x < y$ and $y < z$ then $x < z$
\end{itemize*}

We can take items from the agenda. For each such item, we compare it against the database, and deduce new relations using the above rules. If a newly generated relation conflicts with anything in the agenda or database, then the document is inconsistent. Otherwise, we will move the item from the agenda to the database, and add newly generated relations to the agenda. If we can clear the agenda, then the document is consistent; otherwise, it is not. Whether we add new relations to the top or bottom of the agenda (achieving depth- or breadth-first search, respectively) is irrelevant to the success of the algorithm, though computational performance differences have not been measured.

Our baselines are the results of the \texttt{tlink\_loop} test (Section~\ref{tlink_loop}) and also the results of closure success according to SputLink~\cite{verhagen2005temporal}. This algorithm detected all known inconsistencies in TimeBank, and found one more; full details are later in Section~\ref{consistent_test} A test TimeML corpus is included with CAVaT for verifying that the consistency checker works, which alternative implementations may use for validation. 

\begin{table}
\begin{center}
\caption{Mapping from TimeML relation types to a simple point-based temporal algebra. The TimeML relation is of the form $a$ \textsc{relation} $b$. Where multiple relations are given, all hold. Similar to the table listed in \cite{verhagen2005temporal}.}
\label{tlink_assertions}
\begin{tabular}{ | l | c | }
\hline
\textbf{TimeML relation type} & \textbf{Relation added to agenda} \\
\hline
\textsc{before} & $a_{end} < b_{start}$ \\
\textsc{after} & $b_{end} < a_{start}$ \\
\textsc{iafter} & $b_{end} = a_{start}$ \\
\textsc{ibefore} & $a_{end} = b_{start}$ \\
\textsc{includes} & $a_{start} < b_{start}$, $b_{end} < a_{end}$ \\
\textsc{is\_included} & $b_{start} < a_{start}$, $a_{end} < b_{end}$ \\
\textsc{begins} & $a_{start} = b_{start}$, $a_{end} < b_{end}$ \\
\textsc{begun\_by} & $a_{start} = b_{start}$, $b_{end} < a_{end}$ \\
\textsc{ends} & $a_{end} = b_{end}$, $b_{start} < a_{start}$ \\
\textsc{ended\_by} & $b_{end} = a_{end}$, $a_{start} < b_{start}$ \\
\textsc{simultaneous} & $a_{start} = b_{start}$, $a_{end} = b_{end}$ \\
\textsc{identity} & $a_{start} = b_{start}$, $b_{end} = a_{end}$ \\
\textsc{during} & $a_{start} = b_{start}$, $a_{end} = b_{end}$ \\
\textsc{during\_inv} & $a_{start} = b_{start}$, $a_{end} = b_{end}$ \\
\hline
\end{tabular}
\end{center}
\end{table}

Below is sample output from a consistency check:

\scriptsize
\begin{verbatim}
cavat> check consistent in 3
# Temporal graph consistency checker v1 loaded
# Checking wsj_0927.tml (id 3)
! Inconsistent closure - could not assert 
 (ei2415_2 < ei2414_1)
\end{verbatim}
\normalsize

\subsubsection{Disconnected sub-graph detection}
\label{split_graph}
After inferring a temporal closure~\cite{verhagen2005temporal} of a document, one is usually left with a single interconnected temporal graph, where nodes are TIMEX3s or EVENTs and edges represent TLINKs. However, disconnected groups of links may exist post-closure. This should be brought to the attention of the user; it often suggests that annotating a small number of additional links can greatly increase the amount of data inferable though closure, and that an annotation is incomplete.

CAVaT's sub-graph identification module, \texttt{split\_graph}, works by processing TLINKs from a document sequentially. We maintain a list of sets that will hold interconnected intervals, beginning with an empty list. For each TLINK, we check to see if either of its arguments (which are both intervals) can be found in any set in our list. If one argument can but the other cannot, the new interval is added to the same set as the found interval. If they are both found in the same set, no action is taken. If they are found in different sets, those two sets are merged. If neither TLINK argument can be found anywhere, a new set holding both intervals is created. This process is repeated until all TLINKs have been processed, at which point each set in the list represents an independent sub-graph of connected intervals.

The module will then report statistics about the graph(s) found in the specified document. These include:

\begin{itemize*}
\item Count of sub-graphs, intervals and TLINKs;
\item The number of ``isolated" sub-graphs -- that is, those described by only one temporal link -- and the proportion of intervals/links used to describe all these isolated sub-graphs;
\item Mean and maximum sub-graph size, and the proportion of the document's intervals that are in the largest sub-graph;
\item The entropy of sub-graph sizes, which acts as a ``fracturedness" measure, showing how far the document is from having one single totally connected temporal graph including all TLINKs;
\item The distribution of sub-graph sizes.
\end{itemize*}

Even though sub-graphs are populated by processing the two intervals of a TLINK at the same time, it is possible to have a sub-graph containing just one node, in the case of a TLINK loop (Section~\ref{tlink_loop}). Note that a document containing intervals but no temporal links between them is marked as ``un-fractured", as this check ignores any items not referenced at least once by a temporal link.

Here is sample output from an attempt to identify disconnected sub-graphs:
\scriptsize
\begin{verbatim}
cavat> check split_graph in 3
# Split graph detection v1 loaded
# Checking wsj_0927.tml (id 3)
Subgraphs found: 13 - composed of 69 nodes and linked
 by 65 TLINKS.
Isolated subgraphs, that contain just one TLINK: 5
 (making up 38.5% of all subgraphs / consuming 14.5%
 of all nodes / described by 7.7% of all TLINKs);
Mean graph size 5.3 nodes; largest subgraph (size 35)
 has 50.7% of all nodes.
Entropy of subgraph sizes:  0.448277644573
    2 nodes: ( 5) .....
    3 nodes: ( 4) ....
    4 nodes: ( 3) ...
   35 nodes: ( 1) .
\end{verbatim}
\normalsize

\subsubsection{Superfluous TLINKs}
\label{tlink_loop}
Some TLINKs in TimeML corpora have been specified that associate an event with itself. For example: 

\scriptsize
\texttt{<TLINK lid="l67" relType="IDENTITY" eventInstanceID="\textbf{ei1241}" relatedToEventInstance="\textbf{ei1241}" />}
\normalsize

In this case, the only information conveyed is that ei1241 is identical to itself, making this a redundant TLINK. CAVaT includes a check that will identify TLINKs where both arguments reference the same event instance or event. Although such TLINK \textbf{loops} might be detected by consistency checking, those which specify a SIMULTANEOUS or IDENTITY relation will not.

Below is sample output, showing some superfluous TLINKs:
\scriptsize
\begin{verbatim}
cavat> check tlink_loop in 165 159 143
# TLINK loop checker v1 loaded
# Checking ABC19980304.1830.1636.tml (id 165)
TLINK ID l23 may be a loop (eventID match), type
 INCLUDES, event ei286 / ei288 - check document
 manually
# Checking wsj_1013.tml (id 159)
TLINK ID l107 loops directly (instanceID match), type
 IDENTITY, event ei2495 / ei2495
# Checking wsj_0586.tml (id 143)
TLINK ID l192 loops directly (instanceID match), type
 BEFORE, event ei1404 / ei1404
\end{verbatim}
\normalsize

\subsubsection{Orphaned object details}
\label{orphans}
There is not yet a definition for TimeML annotation completeness, that states a minimal satisfactory level of annotation for a document. In the absence of such a definition, it is not a mistake to annotate entities without attaching them to anything else in the document. However, we believe that wherever possible, every interval should be connected to at least one other interval, and that the annotation of entities that do not contribute or relate to any other annotated information is superfluous. For example, if one chooses to mark text as a temporal signal, a related link or event instance should reference the signal. In this example, if the signal conveys no temporal information, it should not be annotated.

To this end, CAVaT includes a module that is aware of five cases which describe objects attached to nothing else, and reports such \textbf{orphan} objects. Firstly, any TIMEX3 that is not related by any link is deemed to be independent. Also, any event instance (from MAKEINSTANCE) that is not referenced by a link is also orphaned. Next, an EVENT that is never instantiated is unattached, as instantiation is required by current TimeML syntax before EVENTs can be linked to anything else. Instances that come from non-existent or mislabelled EVENTs are also orphans. Finally, SIGNALs that are not referenced by any link or event instance (as in our example above) are included in the list of orphaned objects.

Here is the sample output from a check for orphans:
\scriptsize
\begin{verbatim}
cavat> check orphans in wsj_0927.tml
# Orphaned tag detection v1 loaded
TIMEX3 t104 not in any link
TIMEX3 t131 not in any link
\end{verbatim}
\normalsize

\subsection{Limitations}
CAVaT is currently limited in the number of objects (based on TimeML tags) that it can store for a single corpus. Objects are stored in MySQL tables, and these are limited by the operating system's maximum file size limit. The maximum number of corpora that CAVaT can stored is restricted to the operating system limit of files in a single directory.

\section{Syntax}
\label{syntax}
Here we briefly introduce CAVaT's basic top-level commands, and some of their more useful features. A full specification of CAVaT's syntax is available at \texttt{http://code.google.com/p/cavat}.

\subsection{Corpus manipulation}
Commands for manipulating TimeML corpora within CAVaT begin with \texttt{corpus}. One may view a list of available corpora with \texttt{corpus list}, and use a name from the resulting list to select a corpus for querying or checking with the \texttt{corpus use} command. It is also possible to view any notes attached to the currently selected corpus by using \texttt{corpus info}. Before one can \texttt{use} a corpus, though, a directory of TimeML files must be imported into CAVaT, using \texttt{corpus import}. One may also opt to fold the corpus on import (see Section~\ref{preprocessing}); a note will be attached to the database if this has been done.

\subsection{Querying}
The \texttt{show} command generates reports from the current corpus. Reports focus on one tag type, and give information about its attributes. One can view all values for a tag with ``list" reports, or the distribution of values with ``distribution" reports, or simply see how many instances of that tag use a particular field with ``state" reports.

The general format for report generation is:

\scriptsize
\texttt{show <report type> of <tag> <field> [as <format>]}
\normalsize

From the above example, \texttt{<tag>} corresponds to a TimeML tag, and is one of \texttt{event}, \texttt{instance}, \texttt{timex3}, \texttt{signal}, \texttt{tlink}, \texttt{slink} or \texttt{alink}. As well as the attributes available for each tag, the following extra values for \texttt{<field>} are available:

\begin{itemize*}
\item For TLINKs, \texttt{signaltext} refers to the text enclosed by the start and end tags of an associated signal;
\item For EVENTs, one may reference all the attributes of a MAKEINSTANCE tag too;
\item In TLINKs, SLINKs and ALINKs the arguments are referred to as \texttt{arg1} and \texttt{arg2}, so that the CAVaT user does not have to worry about the implicit indication of interval type present in attribute names.
\end{itemize*}

Reports are available in multiple formats. These can be specified by adding \texttt{as <format>} to the end of a show query.

\begin{itemize*}
\item \texttt{screen} - The default choice, screen gives text formatted for display in a fixed-width font.
\item \texttt{csv} - Output as comma separated values.
\item \texttt{tex} - TeX table format, including caption.
\end{itemize*}

The TeX output of an example report, showing the distribution of TLINK relTypes in TimeBank v1.2, can be generated with \texttt{show distribution of tlink reltype as tex} and is shown in Table~\ref{tab:Tlinkreltype-Frequency-Proportion-distribution}.

\begin{table}
\begin{center}
\caption{Distribution of Tlink reltype}
\label{tab:Tlinkreltype-Frequency-Proportion-distribution}
\begin{tabular}{ | l | r | r | r | }
\hline
\textbf{Tlink reltype} & \textbf{Frequency} & \textbf{Proportion} \\
\hline
BEFORE & 1408 & 21.9\% \\
IS\_INCLUDED & 1357 & 21.1\% \\
AFTER & 897 & 14.0\% \\
IDENTITY & 743 & 11.6\% \\
SIMULTANEOUS & 671 & 10.5\% \\
INCLUDES & 582 & 9.07\% \\
DURING & 302 & 4.71\% \\
ENDED\_BY & 177 & 2.76\% \\
ENDS & 76 & 1.18\% \\
BEGUN\_BY & 70 & 1.09\% \\
BEGINS & 61 & 0.95\% \\
IAFTER & 39 & 0.608\% \\
IBEFORE & 34 & 0.53\% \\
DURING\_INV & 1 & 0.0156\% \\
\hline
Total & 6418 &  \\
\hline
\end{tabular}
\end{center}
\end{table}

One may also specify a subset of a corpus to be used for reporting, using a simple \texttt{where} clause. For example, one may ask:

\scriptsize
\texttt{cavat> show state of tlink signalid where reltype is after} 
\normalsize

to see how many TLINKs of type \textsc{after} use a signal; or, one may ask:

\scriptsize
\texttt{cavat> show distribution of tlink reltype where signalid is not filled}
\normalsize

to find out which relTypes are most likely in TLINKs that do not specify a signal. As part of CAVaT's goal to be easy to use and close to natural language, there are multiple valid syntaxes for filled/unfilled attributes.

\subsection{Browsing}

The ability to examine annotated entities in a TimeML corpus is required as part of investigative research. To enable this, CAVaT includes the \texttt{browse} command.

Browsing allows the user to select a document (with \texttt{browse doc}, followed by a document ID or filename), and then view any tag within that document. Associated data is also shown; for example, if one browses an EVENT tag, any related MAKEINSTANCE tags will also be listed. One may view the tag in three formats -- \texttt{screen}, the default; \texttt{csv}, as two rows of comma-separated values (the first with attribute names as column headings); or \texttt{timeml}, giving valid TimeML for the requested object. The syntax for these is the same as that for \texttt{show} commands; simply append \texttt{as <output type>} to the browse command.

The document selected for browsing is also used as the default document for checks, which are detailed in Section~\ref{checks}

\section{Example tasks}
\label{queries}
Below are some examples of using CAVaT to address real research problems. All are based on TimeBank v1.2.

\subsection{Show all temporal links that employ a signal}
As part of research toward better automatic TLINK annotation, we wanted to know what proportion of TLINKs in a corpus had been annotated as employing a signal.

\scriptsize
\begin{verbatim}
cavat> show state of tlink signalid
  Count  State of Tlink signalid
 ===========================================
    718  signalid filled   (11.2%)
   5700  signalid unfilled (88.8%)
\end{verbatim}
\normalsize

The \texttt{state} keyword here treats signalID as having two states -- filled or unfilled. The TLINK's \texttt{signalid} field will either be empty/absent or contain a reference to a signal annotated in text; for this task, we do not care which specific signal is being referenced.

\subsection{Dealing with ambiguous ``part of speech" values}
Many instances of events in TimeBank assert \texttt{pos="other"}. This is a problem when, e.g., using WordNet to lemmatise event strings. The distribution in Table~\ref{tab:Eventpos-Frequency-Proportion-distribution} can be created with the command:

\scriptsize
\texttt{cavat> show distribution of event pos}
\normalsize
\texttt{ } \\

\begin{table}
\begin{center}
\caption{Distribution of Event part-of-speech}
\label{tab:Eventpos-Frequency-Proportion-distribution}
\begin{tabular}{ | l | r | r | r | }
\hline
\textbf{Event pos} & \textbf{Frequency} & \textbf{Proportion} \\
\hline
VERB & 5122 & 64.5\% \\
NOUN & 2225 & 28.0\% \\
OTHER & 299 & 3.77\% \\
ADJECTIVE & 266 & 3.35\% \\
PREPOSITION & 28 & 0.353\% \\
\hline
Total & 7940 &  \\
\hline
\end{tabular}
\end{center}
\end{table}

After this, we would like to view event text that is classified as \texttt{other}, using the following query:
\scriptsize
\begin{verbatim}
cavat> show list of event text where pos is other
#10.86
#39.8 million
#54.8 million
$1
$1.05
(truncated)
\end{verbatim}
\normalsize

The result suggests that there are at least some numeric values for these event tokens, as well as the more typical verbs. This led to the substitution of all currency and numeric event strings with representative tokens, as a feature for a CRF classifier, yielding a performance increase in TLINK classification (in unpublished results).

\subsection{Which signals does the \textsc{before} relation use?}
Sometimes, particular relation types are strongly suggested by related signals. To determine the signal texts used with \textsc{before} TLINKs, one may query:

\scriptsize
\texttt{cavat> show distribution of tlink signaltext where
 reltype is before}
\normalsize

\begin{table}
\caption{Distribution of Tlink signal text when Reltype is ``before"}
\label{tab:Tlinks.textwhenReltypeisbefore}
\begin{tabular}{ | l | r | r | r | }
\hline
\textbf{Signal text} & \textbf{Frequency} & \textbf{Proportion} \\
\hline
before & 24 & 31.6\% \\
Previously & 10 & 13.2\% \\
by & 7 & 9.21\% \\
already & 6 & 7.89\% \\
Earlier & 6 & 7.89\% \\
until & 5 & 6.58\% \\
then & 4 & 5.26\% \\
followed by & 2 & 2.63\% \\
prior to & 2 & 2.63\% \\
\emph{Other signals, frequency 1} & 10 & 13.2\% \\
\hline
Total & 76 &  \\
\hline
\end{tabular}
\end{table}

From the results in Table~\ref{tab:Tlinks.textwhenReltypeisbefore}, we can see that the token ``before" suggests a \textsc{before} relation, but that the majority of annotated \textsc{before} relations do not employ this signal (from Table~\ref{tab:Tlinkreltype-Frequency-Proportion-distribution}, there are a total of 1408 such relations, only 24 of which use the signal). This indicates that building a relation classifier that relies solely on such signals will not be useful.

\subsection{Superfluous TLINK checking}
One may want to find instances where a link has been made between an entity and itself. We have an error checking module for this, \texttt{tlink\_loop}:

\scriptsize
\begin{verbatim}
cavat> check tlink_loop in WSJ910225-0066.tml
TLINK ID l383 matches, type IS_INCLUDED, event ei1482
TLINK ID l376 matches, type AFTER, event ei1454
TLINK ID l345 matches, type AFTER, event ei1356
\end{verbatim}
\normalsize

One can explicitly query \texttt{in all} to search the entire corpus for similar mis-annotations.

\section{Validation of a sample corpus}
\label{survey}

As we can now load and process any TimeML corpus, and have a set of advanced validation tests, it is logical to test existing TimeML annotated corpora and examine them. In this section, we present the results of running CAVaT's check modules on TimeBank v1.2. This corpus is not new and has been amended and improved by the community~\cite{boguraev2007timebank}, so may contain many fewer errors than freshly annotated documents.

\subsection{Checking for loops}
\label{tlink_loopCheck}
We used the \texttt{tlink\_loop} module (Section~\ref{tlink_loop}) on the corpus. This identifies TLINKs where both arguments are the same event or event instance.

Of TimeBank's 183 documents, 19 have at least one TLINK containing such a loop, and there are 26 loops in total. Of these loops, 10 are on TLINKs of type \textsc{simultaneous} or \textsc{identity}. Such TLINKs will not make a graph inconsistent, but are certainly redundant. The remaining 16 loops of other types will cause an inconsistency. All but one of the loops found are temporal links where both arguments reference the same event instance; only one references two different instances of the same event (TLINK L23, in document ABC19980304.1830.1636.tml). The TimeML in question is as follows:

\scriptsize
\texttt{But they still have <EVENT eid="e28" class="I\_ACTION">catching</EVENT> up to do two hundred and thirty four Americans  have <EVENT eid="\textbf{e30}" class="OCCURRENCE">flown</EVENT> in space, only twenty six of them women.}

\texttt{<MAKEINSTANCE eventID="\textbf{e30}" eiid="\textbf{ei286}" tense="PRESENT" aspect="PERFECTIVE" polarity="POS" cardinality="\underline{234}" pos="VERB"/>}

\texttt{<MAKEINSTANCE eventID="\textbf{e30}" eiid="\textbf{ei288}" tense="PRESENT" aspect="PERFECTIVE" polarity="POS" cardinality="\underline{26}" pos="VERB"/>}

\texttt{<TLINK lid="l23" relType="\textbf{INCLUDES}"
 eventInstanceID="\textbf{ei286}" relatedToEventInstance="\textbf{ei288}"/>}
\normalsize

In this case, the annotation suggests that during the flying in space of 234 Americans, 26 women flew, which is a correct interpretation of the text. CAVaT recommends the manual examination of \texttt{eventID} loops upon their detection. All the other tags reported by this check indicate redundant or incorrect annotations.

\subsection{Checking for consistent graphs}
\label{consistent_test}

Since the consistency checker uses a novel method (see Section~2.3.1), we verified its output by comparing it with that of SputLink and CAVaT's loop detection, and finding explanations for every inconsistency. A small test corpus of TimeML documents is also included with CAVaT for assuring the accuracy of this tool.

SputLink would not report an inconsistency with a TLINK loop that was not of type \textsc{simultaneous} or \textsc{identity}; many of the inconsistent documents were found faulty by both SputLink and CAVaT. Some documents had an erroneous initial TLINK configuration; most faults were subtler than this, and their discovery required a closure attempt.

\subsection{Checking for split graphs}
The \texttt{split\_graph} module checks for documents whose temporal graphs contain sets disconnected TLINKs. No single document in TimeBank has a fully-connected temporal graph, with a path traceable between every interval. The ``best-connected" document (least fractured) is \texttt{wsj\_0144.tml}, which has 34 intervals split into only two subgraphs; one containing 32 intervals, the other two. 

The most fractured document is \texttt{wsj\_1033.tml}, which is split into 12 sub-graphs having a mean graph size of only 2.7 intervals (a single TLINK connected to no other creates a graph of size 2). Despite having 32 intervals in total to connect, the largest sub-graphs in this document include only 4 intervals.

\subsection{Replication}
The results above can be simply replicated by downloading CAVaT v1, gathering a copy of TimeBank v1.2\footnote{LDC catalogue number LDC2006T08}, importing the \texttt{data/timeml/} subdirectory of TimeBank, and running \texttt{check \emph{test} in all} in CAVaT, where \texttt{\emph{test}} is the name of the desired test module.

\section{Conclusion and future work}
\label{conclusion}
We have described CAVaT, a language-independent tool which adds a layer of abstraction between TimeML markup and human researchers, making data easier to analyse, and patterns easier to spot. It also helps identify trouble spots in annotations.

TimeML corpora are only available at this time in Romanian~\cite{for07} and English; this makes multilingual testing of the tool difficult. However, the markup is not language-specific, and results are likely to be equally useful across many languages; this may be shown using test corpora released for TempEval 2\footnote{http://www.timeml.org/tempeval2/}, which will include English, Italian, Spanish, Chinese, Korean and French.

\paragraph{Future work}
CAVaT may be able to provide repair suggestions. These may include fixes for inconsistent graphs, as well as suggestions for missing fields based on lexical resources, third-party tools or heuristics. The modular error checks allow creation of an open database of TimeML validations, to help improve the integrity of all TimeML corpora. Check modules that match the output of rule-based high confidence tools such as S2T~\cite{verhagen2008temporal} can be added.

The consistency checker is ``a TimeML closure engine that uses the precise relations behind the scenes"~\cite{verhagen2005temporal}. Therefore, it may be used to empirically discover how often incorrect links are introduced in closure, when compared with the existing leading closure tool, SputLink.

\paragraph{Acknowledgements}
The first author would like to acknowledge the UK Engineering and Physical Science Research Council for support in the form of a doctoral studentship, and Marc Verhagen of Brandeis University for useful comments on the temporal closure process.

\bibliographystyle{lrec2006}
\bibliography{cavat}
\end{document}